\documentclass{article}

\usepackage[preprint, nonatbib]{nips_2018}

\usepackage{float}

\usepackage[square,numbers]{natbib}
\usepackage[utf8]{inputenc} 
\UseRawInputEncoding
\usepackage[marginal]{footmisc}
\usepackage[T1]{fontenc}    
\usepackage{hyperref}       
\usepackage{url}            
\usepackage{booktabs}       
\usepackage{amsfonts}       
\usepackage{nicefrac}       
\usepackage{microtype}      
\usepackage{graphicx}
\usepackage{epstopdf}
\usepackage{tabularx}
\usepackage{listings}
\usepackage{hyperref}
\usepackage{algorithm}  
\usepackage{algorithmicx}  
\usepackage{algpseudocode}
\usepackage{amsmath} 
\usepackage{times}
\usepackage{color}
\usepackage{tabularray}
\title{Reasoning on Efficient Knowledge Paths:Knowledge Graph Guides Large Language Model for Domain Question Answering}

\author{Boran Jiang\textsuperscript{1}, Yuqi Wang\textsuperscript{1},Yi Luo\textsuperscript{1}, Dawei He\textsuperscript{1} Peng Cheng\textsuperscript{1},Liangcai Gao\textsuperscript{2}\thanks{Corresponding author:gaoliangcai@pku.edu.cn} \\
\textsuperscript{1}State Key Laboratory of Media Convergence Production Technology and Systems, China \\
\textsuperscript{2}Institute of Computer Science and Technology, Peking University, Beijing, China}

\begin{document}

\newcommand{\eraser}{ERASER} 
\newcommand{\yuqi}[1]{\textcolor{blue}{(yuqi) #1}}
\renewcommand{\yuqi}[1]{}
\maketitle
\footnote{First Author and Second Author contribute equally to this work.}
\begin{abstract}
Large language models (LLMs), such as GPT3.5, GPT4 and LLAMA2 perform surprisingly well and outperform human experts on many tasks. However, in many domain-specific evaluations, these LLMs often suffer from hallucination problems due to insufficient training of relevant corpus. Furthermore, fine-tuning large models may face problems such as the LLMs are not open source or the construction of high-quality domain instruction is difficult. Therefore, structured knowledge databases such as knowledge graph can better provide domain background knowledge for LLMs and make full use of the reasoning and analysis capabilities of LLMs. In some previous works, LLM was called multiple times to determine whether the current triplet was suitable for inclusion in the subgraph when retrieving subgraphs through a question. Especially for the question that require a multi-hop reasoning path, frequent calls to LLM will consume a lot of computing power. Moreover, when choosing the reasoning path, LLM will be called once for each step, and if one of the steps is selected incorrectly, it will lead to the accumulation of errors in the following steps. In this paper, we integrated and optimized a pipeline for selecting reasoning paths from KG based on LLM, which can reduce the dependency on LLM. In addition, we propose a simple and effective subgraph retrieval method based on chain of thought (CoT) and page rank which can returns the paths most likely to contain the answer. We conduct experiments on three datasets: GenMedGPT-5k \cite{li2023chatdoctor}, WebQuestions \cite{berant2013semantic}, and CMCQA \cite{CMCQA}. Finally, RoK can demonstrate that using fewer LLM calls can achieve the same results as previous SOTAs models.
\end{abstract}

\section{Introduction}
The advent of large language models (LLMs) \cite{chiang2023vicuna,taori2023alpaca} represents a significant milestone in the realm of pre-trained language models, garnering widespread acclaim. Through extensive unsupervised training on vast datasets, these language models are deemed to have acquired substantial knowledge and intelligence. However, their performance in domain-specific knowledge retrieval, particularly in addressing queries within vertical domains, is suboptimal \cite{bang2023multitask}. Despite their excellent performance, LLMs still have limitations in the following areas: (1) The problem of hallucination. LLMs often generate responses that seem reasonable but are actually wrong \cite{ji2023survey}. Especially in some vertical domains, domain knowledge has a high degree of similarity, and LLMs often confuse questions or answers. This has become one of the important factors hindering the application and promotion of LLMs in the domain. (2) The reasoning ability of complex tasks. LLMs used a large amount of rich corpus during the pre-training stage, and from the actual test, they have a good ability to memorize this knowledge. However, it is often difficult to achieve a satisfactory rate of correctness even when using CoT when faced with tasks that require complex reasoning, because we are not sure whether LLMs have learned the logical relations between knowledge. (3) LLMs lack interpretability. Due to the lack of interpretability in deep neural network inference processes, LLMs are still considered a black box model.

In preceding studies, the Retrieval Augmented Generation (RAG) \cite{lewis2020retrieval} architecture has been employed to address hallucination issues. Researchers try to use the retrieval of question-related documents from external or internal knowledge bases as the basis for LLM’s responses \cite{borgeaud2022improving,shi2023replug}. However, most methods rely on text embeddings to determine semantic similarity, which can lead to the introduction of a large number of semantically irrelevant documents that can affect LLM's understanding of the question. As an unstructured data, documents contain a large amount of redundant information, especially when multiple documents are retrieved, the correlation between documents is sometimes ignored by LLM. On the contrary, Triplets in the Knowledge Graphs (KGs) can represent the relations between entities or events more concisely and efficiently. The application of KGs to enhance LLMs in the context of question answering is a common methodology \cite{pan2024unifying}. Some researchers \cite{wilmot2021memory,li2023chain,jiang2023structgpt} have explored the use of KGs as external knowledge bases to stimulate the reasoning capabilities of LLMs and alleviate the illusion of LLMs. The overall architecture of these methods is to retrieve triples from KGs, and then use them as background knowledge to form prompts to participate into LLM inference. These methods usually perform well on simple questions, while it is difficult to obtain high-quality answers for questions that require multi-hop reasoning in KGs.

\begin{figure*}
    \setlength{\abovecaptionskip}{0.2cm}
    \setlength{\belowcaptionskip}{-0.3cm}
    \centering
    \includegraphics[scale=0.46]{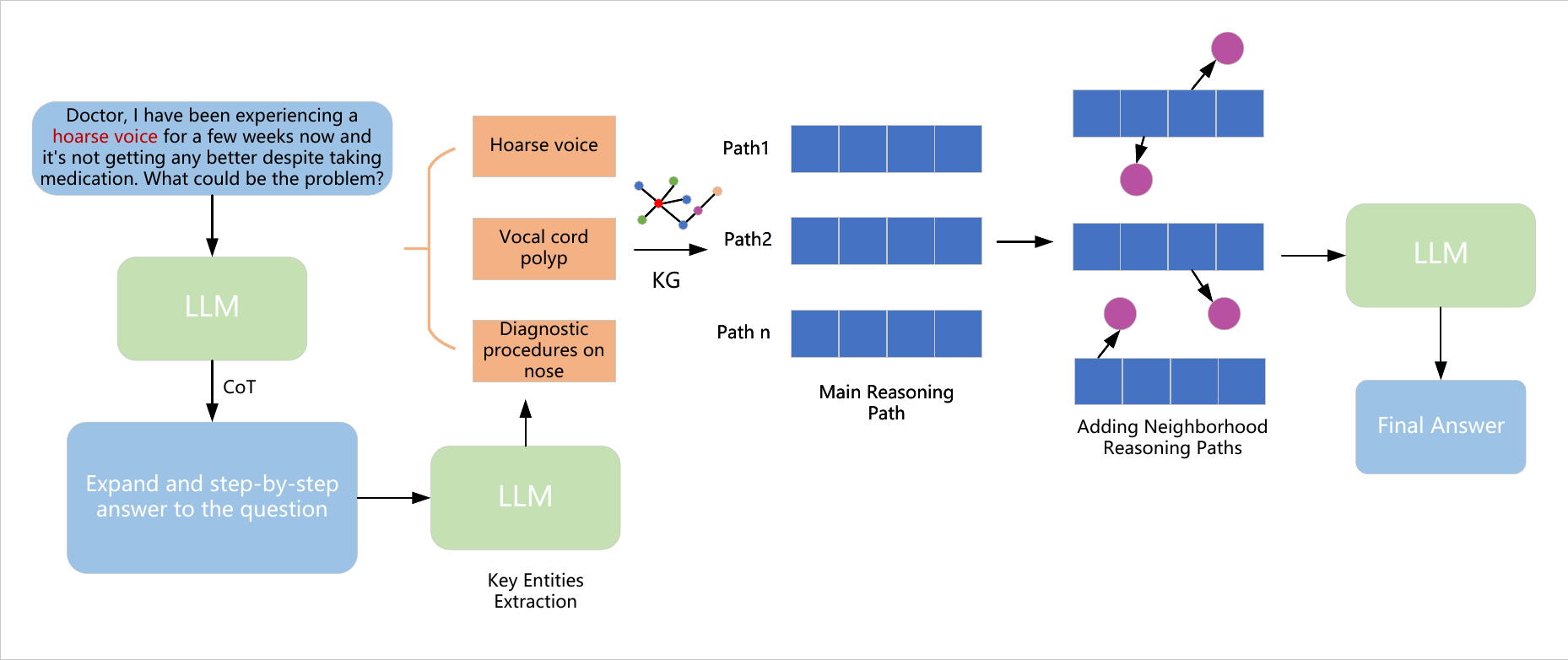}
    \caption{An overview of the architecture of RoK.}
    \label{fig:RoK} 
\end{figure*}

In this paper, we propose a new paradigm, termed Reasoning on Efficient Knowledge Paths (RoK). The RoK can accurately and efficiently select knowledge paths from KG and can fully stimulate the reasoning ability of LLMs. Figure~\ref{fig:RoK} shows the overall framework of RoK. We evaluate our approach on three datasets: GenMedGPT-5k, CMCQA and WebQuestions. The experimental results indicate that RoK can achieve better results with the least number of LLM calls.

\section{Related Work}
\label{sec:related-word}

\textbf{In-Context Learning (ICL) ~\cite{brown2020language}.} is a method of using LLMs for natural language prompts, providing a viable alternative to fine-tuning across various scenarios. The prompts utilized in ICL encompass task descriptions, occasionally supplemented with example sets, serving as inputs to the model to generate desired outputs. ICL facilitates the adaptation of models to diverse tasks without direct parameter modifications. As outlined by \cite{sun2023survey}, ICL can be classified into demonstration example selection, chain-of-thought, and multi-round prompting. 
Despite the remarkable advancements achieved by LLMs facilitated by In-Context Learning (ICL), researchers continue to grapple with the persistent issue of hallucinations. Retrieval Augmented Generation (RAG) is a promising technique that can dynamically incorporate additional evidence into LLM inference \cite{lewis2020retrieval}. A prevalent methodology involves vectorizing relevant documents to construct a related vector database. These methods retrieve the document fragments that are most semantically relevant to the input content through embedding, and then merge them into the prompt \cite{gao2023retrieval}. Recently, LLMs has made breakthrough progress in understanding long contexts \cite{levy2022diverse}, while this vector query approach often fails to accurately match related document chunks and destroys the intrinsic structure between different document chunks in the original document. Moreover, LLMs frequently fail to effectively utilize the retrieval results, and disregarding the central portion of the input and exacerbating hallucination issues \cite{su2022selective}. Consequently, these limitations curtail the applicability of LLMs in precision-critical domains such as healthcare, industry, and productivity tools.

\textbf{KG enhanced LLM.} The knowledge graph stands as a high-precision knowledge repository, displaying notable complementarity with LLMs. Regarding LLM-enhanced graphs, a novel approach involves utilizing LLM to generate the foundational framework of the SPARQL query, while the knowledge graph fills in comprehensive information \cite{li2023few}. Some researchers use LLM to decompose complex problems into distinct sub-problems, subsequently leveraging fine-tuned Llama to generate executable SPARQL queries for retrieving knowledge from KGs \cite{li2023chain}. However, KGs are usually not all imported into LLM, which may result in LLM not being able to write SPARQL correctly from the question description, especially for answers that require multi-hop queries.

In the context of graph-enhanced LLMs, previous research \cite{pan2024unifying} illustrates the multifaceted role knowledge graphs play in the training, prompting, and output phases of LLMs. Notably, the integration of knowledge graphs into the prompting phase has given rise to the Retrieval-Augmented Generation (RAG) architecture based on KGs. However, existing methods either focus solely on KG-related tasks or simply recall retrieved facts, neglecting the underlying structure of the knowledge graph. As LLMs comprehension were deeply explored, researchers extract key entities from the question as seed nodes then use LLMs to select reasoning paths through these nodes on KGs. This helps LLMs to better understand and fully utilize KGs. For instance, MindMap \cite{wen2023mindmap}, beyond choosing multi-hop structures, establishes knowledge paths connecting multiple seed nodes. Think-on-Graph (Sun et al. 2023) constructs a beam search structure, facilitating the search and pruning optimization process on the graph structure by LLMs, aiming to discover the main themes of the query and relationships between these themes. While, these approaches usually rely on the selection of seed entities and require multiple calls to LLMs in the subsequent reasoning path selection, which is not optimistic in terms of operational efficiency.

In summary, our contributions in this paper are as follows:
\begin{enumerate}
  \item We optimize the QA task framework by combining LLMs and KG, which requires at least three calls to LLMs to obtain reasoning paths and high-precision answers, and the dependency on LLMs is much smaller than the current methods.
  \item We introduce the chain of thought to first extend the answer to the question in order to obtain more candidate key entities. The rich candidate entities can increase the probability of matching with entities in the knowledge graph, which ultimately improves the accuracy of path selection.
  \item We propose a new knowledge graph reasoning path selection method that can accurately and comprehensively select the most relevant reasoning path for questioning. 
\end{enumerate}

\section{Problem Definition}
\label{Problem Definition}
Our paper aims to inject domain knowledge into large language model by filtering question-related path information from the knowledge graph. The domain knowledge is represented as a set of triples $G = \{ ({e_s},r,{e_o})|{e_s},{e_o} \in E,r \in R\} $, where ${e_s},{e_o}$ denote subject and object entities and r is a specific relation between two entities. Following prior work (Wen et al., 2023), we use a similar knowledge graph construction method. Given a question \emph{q}, LLM and knowledge graph generate a response by maximizing the probability distribution $p(a|G,{\phi _{LLM}},q)$ where ${\phi _{LLM}}$ represents the parameters of the large language model (LLM). However, limited by the length of input context of LLM, we are unable to utilize the entire knowledge graph to construct prompts. To address this challenge, we extract the entity ${e_q}$ mention in question through a large language model (LLM) and then link entities ${E_g} \in E$ in knowledge graph G. The subgraph ${G_{sub}} \in G$ containing answers to the question can be generated through the path between $e \in {E_g}$. Finally, the most relevant knowledge paths ${p_{sub}}$ can be further filtered through the subgraph ${G_{sub}}$. We can rewrite $p(a|G,{\phi _{LLM}},q)$ as:

\begin{equation}\label{eqn-1} 
  p(a|G,{\phi _{LLM}},q) = p(a|{p_{sub}},{\phi _{LLM}})p({p_{sub}}|q,{G_{sub}})
\end{equation}

\section{Methods}
\label{sec:methods}
RoK selects the path most relevant to the input question on the knowledge graph. Specifically, RoK can expand the question or query \emph{q} through the chain of thought and then find more related entities. These entities can be better aligned with our pre-constructed knowledge graph from multiple perspectives. In many cases, the input question semantics are concise or use a variety of descriptors, which will lead to failure to match entities with the knowledge graph. To address this problem, we use CoT to obtain a richer entity set ${E_q}$ from the question and link it to the knowledge graph to obtain the aligned entity set ${E_g} = \{ e_g^1,e_g^2,...,e_g^k\}$. RoK obtains the subgraph ${G_{sub}}$ through path matching between entity nodes and uses PageRank to filter out the top-N reasoning paths for the question. These paths serve as domain knowledge prompts to input into the large language model and help it obtain the accurate answer.
We show the main framework in Figure~\ref{fig:RoK}, which comprises two main components:
\begin{enumerate}
    \item \textbf{LLM generates reasoning steps for query}: RoK leverages the CoT of LLM to expand the question and derive the answer step by step, with the purpose of obtaining the entities in the reasoning step. The rich entities are helpful for subsequent selection of more detailed knowledge paths from the knowledge graph. 
    \item \textbf{Knowledge reasoning path generation}: Select candidate paths containing the most useful background knowledge by traversing multi-hop paths between key entities. These background knowledge paths can prompt the LLM to find the correct answer to the query.
\end{enumerate}

\subsection{LLM generates reasoning steps for query}
\begin{figure*}
  \setlength{\abovecaptionskip}{0.2cm}
  \setlength{\belowcaptionskip}{-0.3cm}
  \centering
  \includegraphics[scale=0.46]{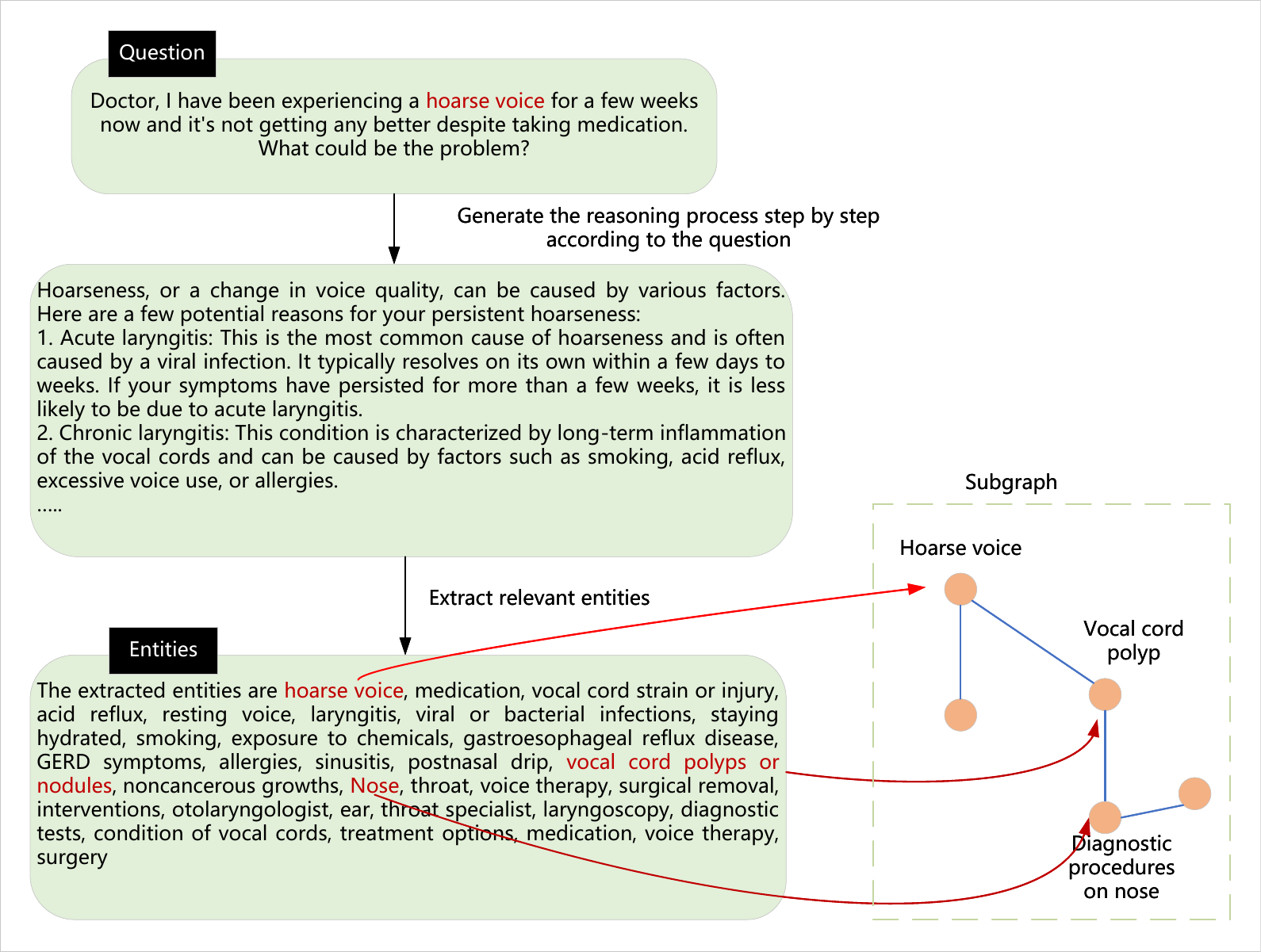}
  \caption{An example workflow of generating reasoning process step by step related to the input problem. Then we leverage LLM to extract key entities from the reasoning process and link to the knowledge graph.}
  \label{fig:Extended CoT} 
\end{figure*}

When using the knowledge graph to assist LLM in answering domain questions, we first need to link the key entities in the question to the knowledge graph. The input question usually contains key entities which can be linked to the knowledge graph. These entities can point to the answer entity in KG through multi hop paths which can be called reasoning paths. Selecting the reasoning path requires first extracting key entities from the question and linking them to the KG. While at this step, there are often have the following issues: (1) The entities in the question cannot be linked to the knowledge graph through text embedding. This may be caused by the lack of relevant training corpus for the text embedding model. (2) Only one key entity in the question can be linked to KG. In this case, the reasoning path just can be generated using a random walk or multiple calls to the LLM to determine the path generation direction \cite{sun2023survey}.
To address these problems, we first use the CoT of LLM to provide extension or reasoning answers to the questions step by step. And our main focus is on the thinking process in answering, as well as the key entities that arise related to the question. Although LLMs are not necessarily good at answering questions in domain fields, their rich corpus in the pre-training stage is likely to contain rich domain knowledge. Therefore, the generated reasoning steps or solution process usually contain a large number of key entities, which can be linked to the knowledge graph. A case can be shown in Figure~\ref{fig:Extended CoT}.

\subsection{Knowledge reasoning path generation}
Generating reasoning paths are divided into two main steps. The first step is to generate the main reasoning paths through key entities and the second step is to generate the neighbor branching paths.The overall reasoning paths selection algorithm of RoK is illustrated in Algorithm 1. The algorithm consists of two parts: main reasoning paths selection and neighborhood branch reasoning paths selection.
\subsubsection{Main Reasoning Path Generation}
\begin{figure*}
    \setlength{\abovecaptionskip}{0.2cm}
    \setlength{\belowcaptionskip}{-0.3cm}
    \centering
    \includegraphics[scale=0.46]{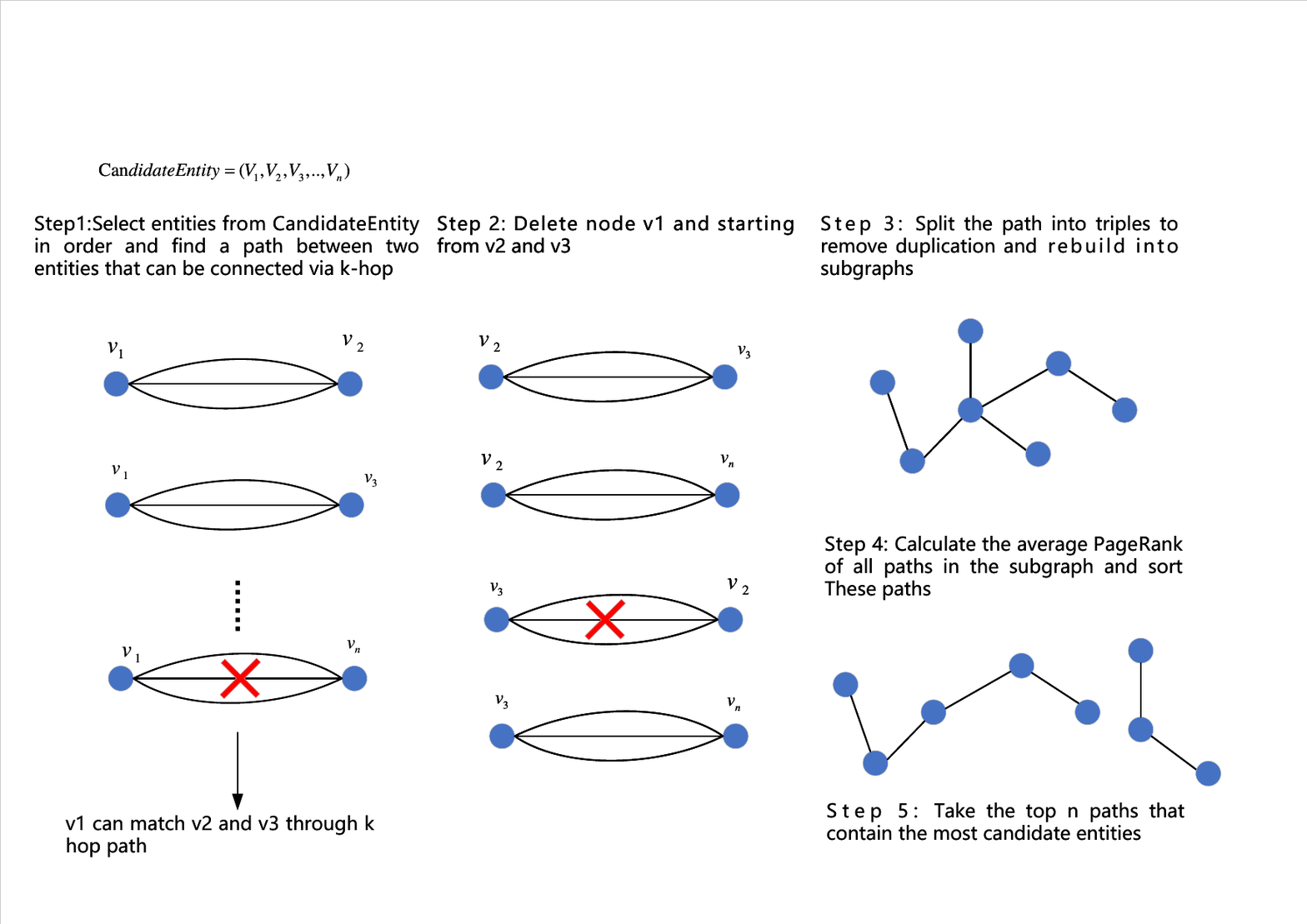}
    \caption{The generation process of the main reasoning path.}
    \label{fig:Main Path} 
  \end{figure*}

We define the entities linked to the knowledge graph as entity candidate sets ${E_{cand}} = \{ {e_1},{e_2},...,{e_n}|e \in G\} $. In this stage, we construct candidate entities in pairs and record the n-hop paths between them to obtain candidate sub-graphs. However, there are still some noisy entities in the subgraph that are not suitable as prompt feed into LLM, we use PageRank to further filter the reasoning paths.
\newline \textbf{Reasoning sub-graph Generation} In section 4.1, we expand the number of key entities by providing initial answers to questions through CoT. These key entities can be used as a summary description of the entire question and they are generally connected within n hops in KG. Therefore, we use the entities in the candidate set as the starting point and the ending point respectively to search for possible paths which may contain the answer entities. We take out two entities ${v_1}$ and ${v_2}$ from the candidate set in order each time and save all n-hop paths between them in ${p_{cand}}$. Then ${v_2}$ was removed from the candidate set of key entities. Traverse the n-hop path between entity ${v_1}$ and all remaining entities, as shown in Figure~\ref{fig:Main Path}, ${v_1}$ has paths with ${v_2}$ and ${v_3}$ in n-hop. In the next iteration, ${v_2}$ and ${v_3}$ will be used as the start node. In this process, there may be some duplicate triples, so we need to deduplicate the triples before each iteration. Finally, these triples can be combined to form a subgraph ${G_{sub}}$.
\newline \textbf{Selecting reasoning paths from subgraph based on PageRank} aims to remain the triples which are more relevant to the question. The sub-graph ${G_{sub}}$ contain the key entities which are highly relevant to the answer. While, there are also noisy entities and relations in the sub-graph. Therefore, we filter out top N reasoning paths that are more relevant to the question. Selecting reasoning paths includes three steps:(a) We calculate the \emph{PR} value of the entity node in the subgraph as $P{R_{({e_i})}}$ through the PageRank algorithm. The specific calculation process is as shown in following function. Where $d(0 \le d \le 1)$ is the damping factor, n is the number of entity nodes, M is the transition matrix which is the transition probability from a node to all nodes connected to it. The second term in function (3) ensures that the PR value of all nodes is not zero.

  \begin{equation}\label{eqn-2} 
    PR({e_i}) = d(\sum\limits_{{e_j} \in {G_{sub}}} {\frac{{PR({e_j})}}{{L({e_j})}}} ) + \frac{{1 - d}}{n},i = 1,2,...,n
  \end{equation}
  \newline By introducing the transition matrix, the upper function can be rewritten as:
  \begin{equation}\label{eqn-3} 
    P{R_t} = dMP{R_{t - 1}} + \frac{{1 - d}}{n}1
  \end{equation}
  \newline (b) Sorting the number of key entities $e \in {E_{cand}}$ contained in the path ${P_{cand}}$ and calculate the average \emph{PR} value for each path. Then we divide the sorted paths into buckets according to the number of key entities they contain. The range of buckets is $(0,1,...,n)$, where \emph{n} is the number of key entities. 
  \newline (c) We select the top \emph{k} paths with the largest average \emph{PR} value from the largest bucket. If ${k<n}$, the path selection continues based on the \emph{PR} value of the second bucket until ${k=n}$. The final set of all reasoning paths is represented as ${P_{main}} = \{ {p_i}|{p_i} \in {P_{cand}},i = 1,2,...,N\}$

\subsubsection{Neighbor branch reasoning path selection}
\begin{figure*}
    \setlength{\abovecaptionskip}{0.2cm}
    \setlength{\belowcaptionskip}{-0.3cm}
    \centering
    \includegraphics[scale=0.55]{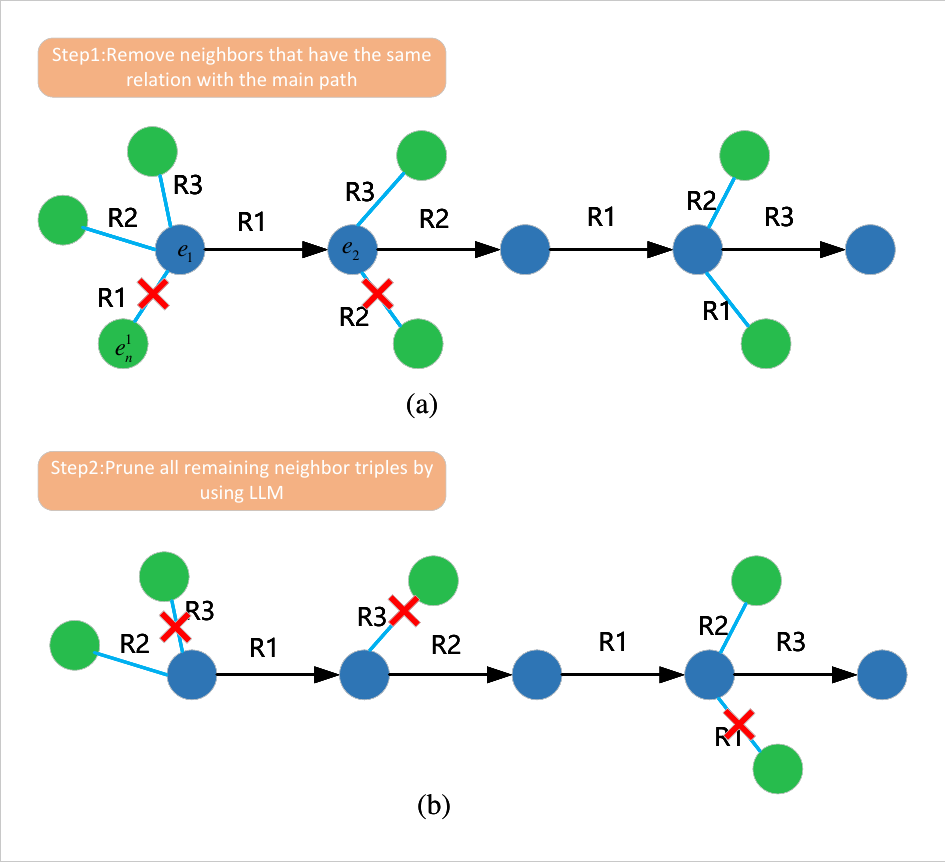}
    \caption{The process of selecting neighbor triple in main reasoning path. The pruning process is divided into two steps, which are removing repeat triples based on same relation and using LLM to select question-related triples.}
    \label{fig:Neighbor Path} 
  \end{figure*}

The main reasoning path always focuses on the explicit answer in the question but may not respond well to the implicit answer in the question. As a supplementary to main reasoning path, we introduce the first-order neighbors of entities on the main path as branch of the reasoning paths. As shown in Figure~\ref{fig:Neighbor Path}, it has two steps: (a) Get the first-order neighbor nodes ${E_N} = \left\{ {{e_i} \in G|{e_i} \notin {P_{main}},i = 1,2,...n} \right\}$. Iterate over every key entity in ${E_{cand}}$, take Figure~\ref{fig:Main Path} as an example, ${e_1}$ is connected to two entities ${e_2}$ and $e_n^1$ through the same relation ${R_1}$. In the same relation, we first remain the entity on the main path and delete the neighbor entity $e_n^1$. (b) Then we will check if all remaining neighbor triples are semantically relevant to the question by using LLM. We observed that previous related work of knowledge graph retrieval-augmented LLMs often only used the relations to judge their relevance to the question when searching paths. Due to missing head and tail entities, the relations often contain incomplete semantic information. Judging semantic relevance to the question only by the relation is usually inaccurate. In this work, we use the entire triplet to determine its semantic relevance to the question. The structure of triples is semantically more coherent. In this step of the process of filtering neighbor entities, all triples are fed into LLM at once, therefore the LLM only need to be called once.
  
\subsection{LLM generates answer by combining reasoning paths}
To get accurate answer, the two reasoning paths are treated as background knowledge used to compose the prompt for LLM. Usually, the main reasoning path contains the entities most relevant to the question and the neighbor branch reasoning path contains descriptions or supplementary information about key entities. The combination of the two paths allows for more comprehensive background information to be included in the prompt. All these prompt templates are shown in Appendix.
\newline LLMs use a rich corpus and knowledge in the pre-training stage. Since knowledge and reasoning are not decoupled during the LLMs training process, we are not sure whether LLMs only remember all the knowledge or fully understand the logical connections behind the knowledge. It has been proven in many related works that even if domain data fine-tuning is used, LLM still suffers from illusion problem that cannot be ignored when answering domain questions. The reasoning path extracted from the knowledge graph contains both knowledge and reasoning process, so it can trigger and guide LLM to generate the final correct answer. And the constraints on the path of reasoning through the graph can effectively mitigate the illusion problem that exists in LLMs.

\begin{algorithm}
  \caption{Optimal Reasoning Paths Selection}\label{Optimal reasoning paths}
      \begin{algorithmic}[1]
\Require Entities matched from the question $E q=\left\{e_{q 1}, e_{q 2}, \ldots, e_{q n}\right\}$; question prompt $q_p$; max\_hop=N; top\_rel=M; top\_paths=k
\Ensure Reasoning paths
\Function{FindNodesPath}{$e,E q$}
  \State path =Find all paths between entity $e$ and other entity in the candidate\_list within max\_hop
  \State match\_entities=Record nodes in paths except for $e$
  \State \textbf{return} $paths,match\_entities$
\EndFunction
\Function{Gen\_main\_reasoning\_paths}{$E q$}
  \State $e\gets candidate\_list[0]$
  \State $all\_path\gets []$
  \While{$E q$ is not empty}
    \State $paths,match\_entities$=FindNodesPath($e,E q$)
    \State $E q$.remove($e$)
    \State $e$=match\_entities
    \State all\_paths.append(paths) 
  \EndWhile
  \State MainReasoningPaths=BucketSelPath($all\_paths$,$top\_paths$) \Comment{Select the top k paths using the average PageRank value of the path and the number of key entities they contains}
  \State \textbf{return} $MainReasoningPaths$
\EndFunction
\Function{Gen\_neighbor\_paths}{$E_q,all\_paths$}
  \State $final\_rdf\gets []$
  \For{$e$ in $E_q$}
    \State rdfs=find\_one\_hop\_path($e$)  \Comment{Finds the first-order neighborhood triplet of entity $e$}
    \For{$rdf$ in $rdfs$}
      \If{$rdf$ not in $all\_paths$ and LLM determines the the rdf is related to the question}
        \State final\_rdf.append($rdf$)
      \EndIf
    \EndFor
  \EndFor
  \State \textbf{return} $final\_rdf$
\EndFunction
      \end{algorithmic}
\end{algorithm}

\section{Experiments}

\begin{table}[t]
\centering
\caption{The statistics and usage details of three datasets.}\label{tab:dataset}
\setlength{\extrarowheight}{2pt}
\resizebox{1\linewidth}{!}{
\begin{tabular}{lccccc}
\specialrule{2pt}{0pt}{0pt}
Dataset      & Domain                                   & Question & Answer type               & Relationship & Entity \\ \hline
GenMedGPT-5k & Diagnose through interrogation (English) & 516      & Disease, Medication, Test & 6            & 1122   \\
WebQuestions & Open-domain from Web (English)           & 400      & Entities                  & 237          & 14951  \\
CMCQA        & Diagnose through interrogation (Chinese) & 400      & Disease, Medication, Test & 12           & 62282  \\
\specialrule{2pt}{0pt}{0pt}
\end{tabular}}
\end{table} 

\begin{table}[t]
\centering
\caption{The BERTScore and GPT-4 ranking of all methods for GenMedGPT-5k and CMCQA.}\label{tab:BERTScore and GPT-4 ranking}
\setlength{\extrarowheight}{2pt}
\resizebox{1\linewidth}{!}{
\begin{tabular}{lccccc}
\specialrule{2pt}{0pt}{0pt}
& BERT Score         &        &                & GPT-4 average Ranking \\    \hline
~                   & Precision          & Recall & F1 Score       &                       \\  \hline
GPT-3.5             & GenMedGPT-5k:0.772 & 0.811  & 0.791          & 3.44                  \\
                    & CMCQA:0.916        & 0.913  & 0.915          & 3.37                  \\  \hline
Embedding retrieval & 0.778              & 0.816  & 0.797          & 3.07                  \\
                    & 0.922              & 0.919  & 0.921          & 3.11                  \\  \hline
MindMap             & 0.798              & 0.806  & 0.802          & 1.94                  \\
                    & 0.939              & 0.936  & 0.937          & 1.76                  \\  \hline
RoK(ours)           & \textbf{0.817}     & 0.813  & \textbf{0.815} & \textbf{1.53}         \\
                    & \textbf{0.941}     & 0.929  & 0.934          & 1.78                  \\               
\specialrule{2pt}{0pt}{0pt}
\end{tabular}}
\end{table}



\begin{table}[t]
\centering
\caption{The key entities average match accuracy for GenMedGPT-5k and CMCQA.}\label{tab:accuracy}
\setlength{\extrarowheight}{1pt}
\resizebox{0.6\linewidth}{!}{
\begin{tabular}{lccccc}
\specialrule{2pt}{0pt}{0pt}
Methods             & Key Entities Average Match Accuracy \\  \hline
GPT-3.5             & GenMedGPT-5k:52.6                   \\   
                    & CMCQA:85.6                          \\
Embedding retrieval & 73.2                                \\
                    & 88.4                                \\
MindMap             & 75.5                                \\
                    & 91.0                                \\
RoK                 & 81.3                                \\
                    & 91.4                                \\
\specialrule{2pt}{0pt}{0pt}
\end{tabular}}
\end{table}

\begin{table}[t]
\centering
\caption{The key entities match accuracy for WebQuestions.}\label{tab:WebQuestions}
\setlength{\extrarowheight}{0.8pt}
\resizebox{0.45\linewidth}{!}{
\begin{tabular}{lccccc}
\specialrule{2pt}{0pt}{0pt}
Methods             & Accuracy      \\  \hline
GPT-3.5             & 68.5          \\
Embedding retrieval & 76.0          \\
MindMap             & 79.8          \\
RoK(ours)           & \textbf{80.5}  \\
\specialrule{2pt}{0pt}{0pt}
\end{tabular}}
\end{table}

In this section, we evaluate our approach on three question-answering datasets: GenMedGPT-5k \cite{li2023chatdoctor}, WebQuestions \cite{berant2013semantic}, and CMCQA \cite{CMCQA}. We design the experiments to mainly answer the three questions: (a) Whether the LLM enhanced by knowledge graph reasoning paths is superior to the vanilla LLMs? (b) Does RoK perform better in QA than previous retrieval-augmented LLMs? (c) Does RoK can choose the most relevant and correct reasoning paths for the question.
\subsection{Experiment Settings}
\subsubsection{Datasets}
\textbf{GenMedGPT-5k} Generate conversations between patients and doctors using ChatGPT and disease databases. We selected 516 QA pairs from GenMedGPT-5k.
\newline \textbf{WebQuestions} contains 6,642 QA pairs and the most questions are popular ones asked on the website. The questions are supposed to be answerable by Freebase \cite{bollacker2008freebase}. Freebase is a large knowledge graph and contains many triples that are not related to WebQuestions, therefore we chose FB15k-237 as the knowledge base. The knowledge base of FB15k-237 is a subgraph composed of a small number of subject words taken from Freebase.
\newline \textbf{CMCQA} is a huge conversational QA dataset in Chinese medical field. It comes from the Chinese Medical Conversation QA website, which has medical conversation materials in 45 departments, including male, otolaryngology, obstetrics and gynecology. Specifically, CMCQA has 1.3 million complete conversations and 650 million tokens.
\newline To construct the knowledge graph for the two datasets GenMedGPT-5k and CMCQA, we referred to the method in MindMap \cite{wen2023mindmap}. The statistics and usage details of three datasets are shown in Table \ref{tab:dataset}
\subsubsection{Evaluation Metrics}
For GenMedGPT-5k and CMCQA, we used BERTScore \cite{zhang2019ernie} to measure the semantic similarity between the generated answer and the ground truth. GPT-4 is used to evaluate the accuracy of the three parts of the answer: Disease (Diagnose possible diseases), Medication (Recommended medication), and Test (Recommend relevant tests based on possible disease). We used two methods to evaluate three-part answers: (a) GPT-4 ranking. Specifically, we asked GPT-4 to rank the answers obtained by different methods according to the correctness. Then we calculate the average value based on the ranking of these answers. The lower the score, the closer the answer is to the reference answer. (b) Key entities average match accuracy. We calculate the average entity hit rate for Disease, Medication and Test separately.
\newline For WebQuestions, we use accuracy (Hits@1) as evaluation metric, because most of the answers are single entity or multiple entities.

\subsection{Main Results}
\subsubsection{Comparison to Other Methods}
We compare with different QA methods based on LLM, including GPT-3.5, MindMap and Document embedding retrieval. Table \ref{tab:BERTScore and GPT-4 ranking} presents the BERTScore and GPT-4 ranking of GenMedGPT-5k and CMCQA these two datasets. BERTScore mainly measures the semantic similarity between the generated answer and reference answer. Since the generated answer is a sentence rather than an entity, its length may affect the final result. It can be seen from Table 2 that the BERTScore of these methods are relatively close. Since the final generated results are dependent on the LLM, there will be some similar template words in the answers. To better evaluate the generated answers, we use GPT-4 average Ranking and key entities average match accuracy. Specifically, we leveraged GPT-4 to generate an accuracy ranking of the answers for the four methods and then calculated the average Ranking. In the GenMedGPT-5k dataset, The GPT-4 average Ranking of RoK shows a significant improvement compared to the vanilla baseline GPT-3.5. Compared with other retrieval enhancement methods the GPT-3.5 had the worst answer accuracy. As can be seen in , most of the medical knowledge in the GPT 3.5 responses was correct, but the knowledge was not well established logically thus leading to a final incorrect answer. This demonstrates that LLMs may have a sufficiently rich corpus of domain knowledge and they can memorize the knowledge well, but they do not sufficiently learn the logical relations between the knowledge.
\newline MindMap, also belonging to retrieval enhancement methods, performs much better than embedded retrieval. This demonstrates that compared to using large sections of documents as prompts, structured reasoning paths as prompts can better stimulate the understanding and reasoning abilities of LLMs.
\newline Since using GPT-4 as a discriminator is not the most objective evaluation method, we introduce a new metric approach. As we can see in Figure~\ref{fig:RoK MindMap reasoning paths} and Figure~\ref{fig:all answers}, the accuracy of the answer mainly depends on the key entities. Therefore, we also measured the key entity matching accuracy as shown in Table 3. It can be seen from Table 3, RoK achieved the best performance. By further analyzing the results, we find that the obtained reasoning path contains most of the correct entities. MindMap also achieves good performance, but the effect decays significantly when it is unable to link to entities from the knowledge graph or extract entities from the question. RoK can effectively mitigate this problem by extending key entities.
\newline Table 4 shows the comparison results for WebQuestions, it can be seen that the performance of the last three methods is not significantly different. Because most questions based on this dataset do not require complex reasoning, ordinary embdding retrieval can also perform well. And RoK is better suited for reasoned answers to complex questions.

\subsubsection{Extensibility of Key Entities}
As shown in Figure~\ref{fig:Extended CoT}, we use CoT to get an initial answer to the question through the vanilla LLM to be able to capture a portion of the key entities related to the question. Even if the final answer obtained is incorrect, it does not affect the expansion of the key entities. Matching to more key entities means we can derive richer background knowledge and reasoning paths from KG. From Figure~\ref{fig:Extended CoT}, it can be seen that only the entity \textit{'house voice'} can be matched in the question. It is difficult to obtain accurate reasoning paths from KG with only one key entity. Expanding to more key entities can better guide the selection of reasoning paths. We show the final answer result in Figure~\ref{fig:RoK MindMap answers} in Appendix, from this case, it can be seen that RoK provided the correct answer. While, MindMap, which also uses the same method of obtaining reasoning paths in KG, generates incorrect answer due to matching too few entities.
\subsubsection{Accuracy analysis of main reasoning path selection}
As shown in Figure~\ref{fig:RoK MindMap reasoning paths} in Appendix, the highlighted red section in the reference answer is the key entity of the answer and the blue color represents incorrect answer. By comparing the selected reasoning paths, it can be found that the starting key entities are all \textit{'headache'}, but the paths selected are quite different. And the reasoning paths have a direct impact on the final answer result. In RoK, we get the reasoning paths by iterating over all the pairwise multi hop paths between all key entities. This method ensures that potential connections between key entities can be fully explored.

\section{Conclusion}
In this paper, we propose an effective method for solving QA problems based on knowledge-based reasoning paths combined with LLM, which can help LLM improve performance on domain-specific knowledge required tasks. By optimizing the generation method of the reasoning path, the final result can be obtained by calling LLM only three times. By introducing external knowledge, the hallucination issue of LLM can be effectively mitigated. Through experiments on three datasets, we demonstrate that our method performs better than the vanilla LLMs and other retrieval-enhanced LLM generation methods.
\clearpage

\bibliography{main}
\clearpage

\appendix
\begin{appendix}
\section{Appendix}

\begin{figure*}[!thbp]
\setlength{\abovecaptionskip}{0.2cm}
\setlength{\belowcaptionskip}{-0.3cm}
\centering
\includegraphics[scale=0.46]{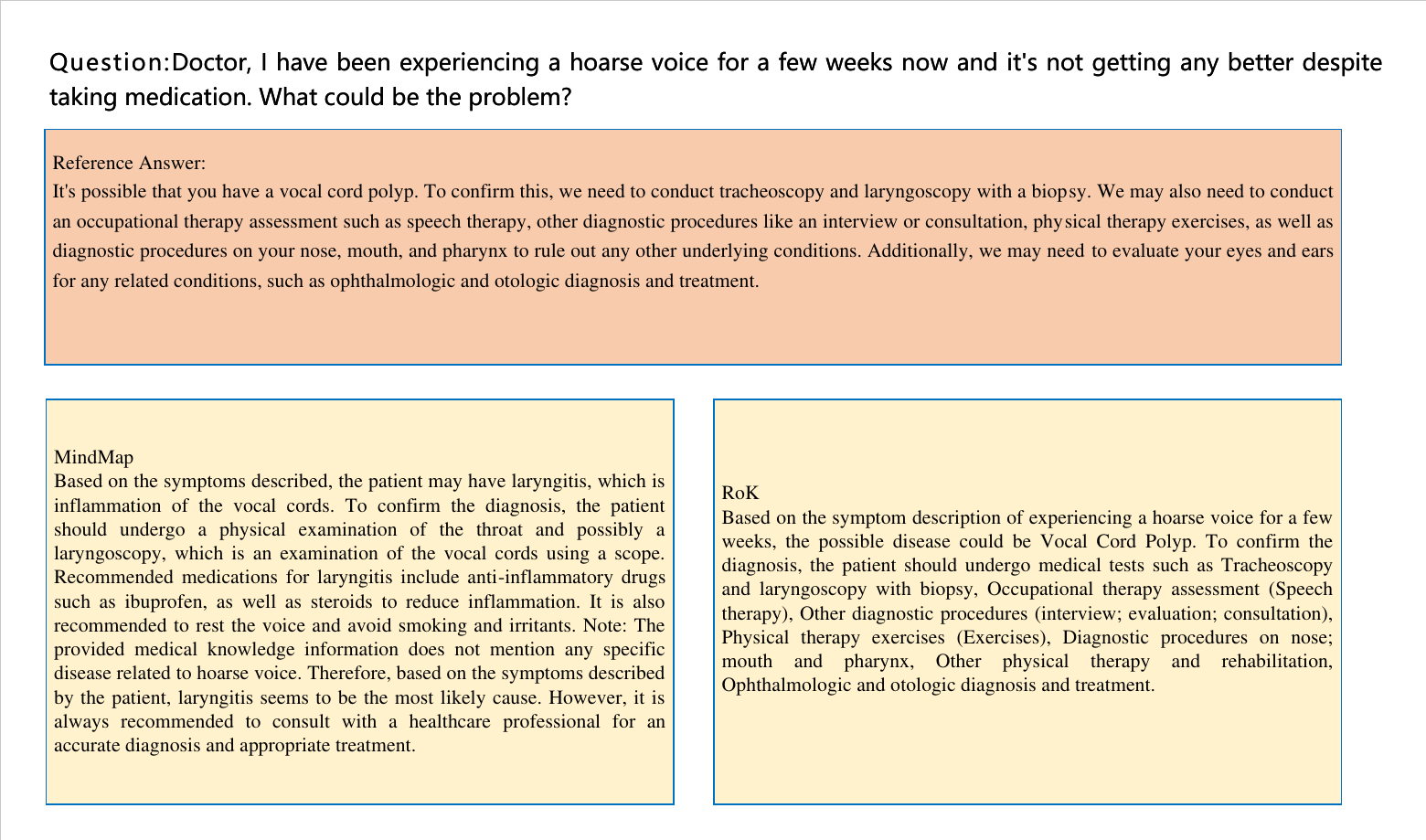}
\caption{Comparison of RoK and MindMap's final answers.}
\label{fig:RoK MindMap answers} 
\end{figure*}

\begin{figure*}[!thbp]
\setlength{\abovecaptionskip}{0.2cm}
\setlength{\belowcaptionskip}{-0.3cm}
\centering
\includegraphics[scale=0.46]{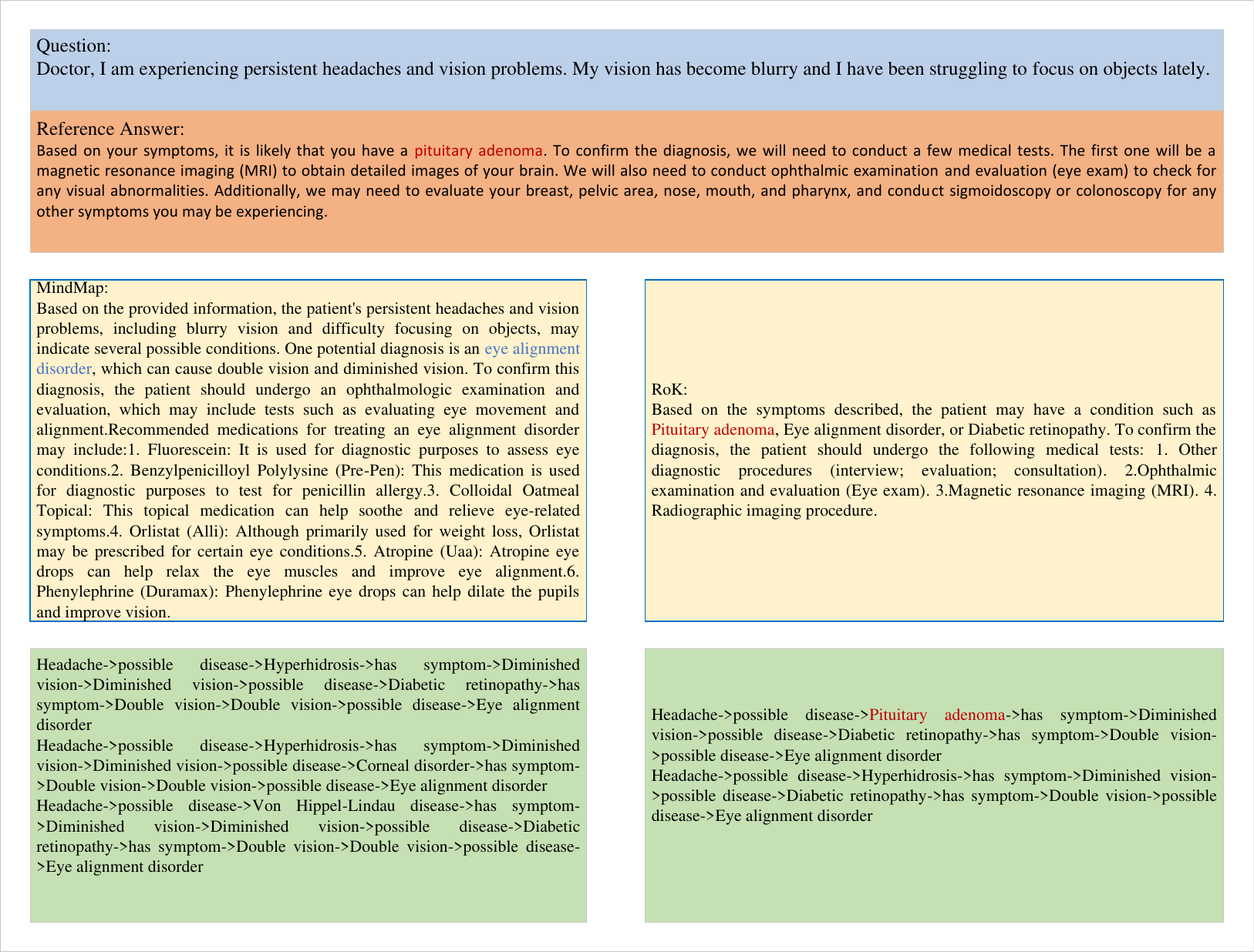}
\caption{A case of comparison between MindMap and RoK approaches on reasoning paths. The right answer is highlighted with red color, and the wrong answer is highlighted with blue color.}
\label{fig:RoK MindMap reasoning paths} 
\end{figure*}

\begin{figure*}[!thbp]
\setlength{\abovecaptionskip}{0.2cm}
\setlength{\belowcaptionskip}{-0.3cm}
\centering
\includegraphics[scale=0.46]{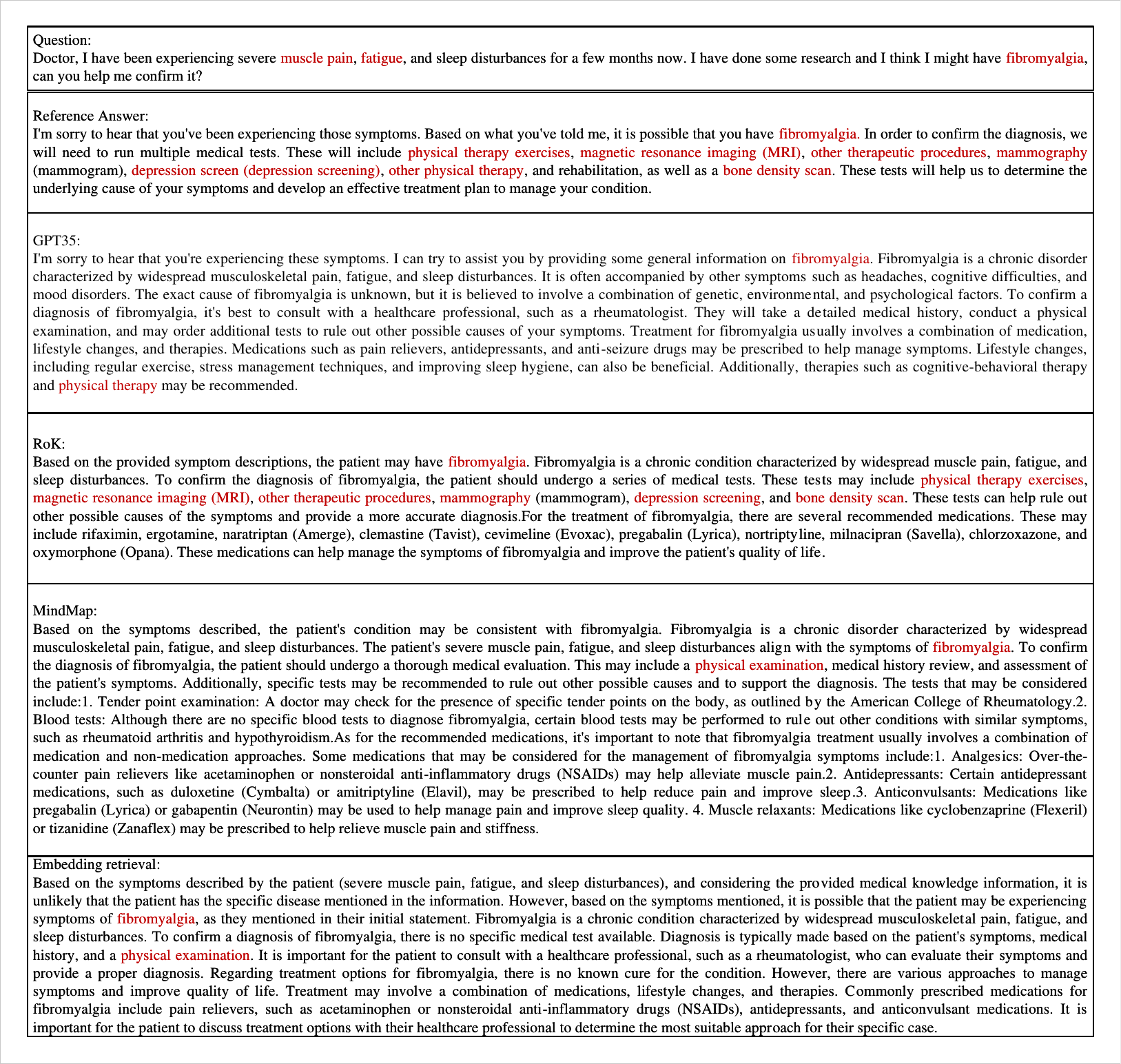}
\caption{A case of comparison of final answers for all methods. The highlighted red area indicates the key entities that have been matched. And these key entities typically contain the correct answer.}
\label{fig:all answers} 
\end{figure*}

\subsection{The Comparison of Experimental Cases}
In this section, we presented the comparison of the answer results between RoK and other methods.
\subsection{The Prompts Used for Experiments}
In this section, we show all the prompts to be used in the experiments. The first prompt initializes an extended answer to the question by introducing the chain of thought, which does not use background knowledge from the knowledge graph. The second prompt extracts key entities from the extended responses as anchor entities for later obtaining the reasoning path from the knowledge graph. The third prompt is used to filter out neighborhood triples related to the question as part of the background knowledge. The last prompt receives the reasoning path and gives the final answer.
\subsubsection{Extending entities through the chain of thought}

\centering
\begin{lstlisting}[language=Python, basicstyle = \ttfamily, breaklines = true]
Template="""
    You are an excellent AI doctor, and you can give disease diagnosis suggestions and analysis process and recommend medications step by step based on the patient's question.
    Patient's question: {}
    Output:
    """
\end{lstlisting}

\subsubsection{Key entity extraction related to question}

\begin{lstlisting}[language=Python, basicstyle = \ttfamily, breaklines = true]
Template="""
    extract the key entities for the following text: {}
    The types of key entities that need to be extracted are related to disease diagnosis, treatment protocols, medications, tests that need to be done, possible disease names, etc
    Output:
    """
\end{lstlisting}

\subsubsection{Filter neighborhood triples}
\begin{lstlisting}[language=Python, basicstyle = \ttfamily, breaklines = true]
Template="""
    Please select the triplets related to the patient's question and the content that needs to be answered. Please refer to the following background knowledge when answering.
    Patent’s question:
    ###{}
    The content that needs to be answered:
    1.What disease does the patient have?
    2.What tests should patient take to confirm the diagnosis?
    3. What recommended medications can cure the disease?
    background knowledge:
    ###{}
    Triplets:
    ###{}
    Output:
    In-Context One-shot
"""
\end{lstlisting}  

\subsubsection{Final Answer}
\begin{lstlisting}[language=Python, basicstyle = \ttfamily, breaklines = true]
Template="""
    You are an excellent AI doctor, and you can diagnose diseases and recommend medications based on the symptoms in the conversation. 
    Patient input: {}
    You have some medical knowledge information in the following:
    ###{}
    ###{}
    The final answer consists of three parts:
    1.What disease does the patient have? If it is not possible to determine from the MEDICAL background knowledge given above what disease the patient is suffering from then this section can be left unanswered or the patient can be referred for tests to determine what disease he may have.
    2.What tests should patient take to confirm the diagnosis? 
    3.What recommended medications can cure the disease? Think step by step.
    Output: The answer includes disease and tests and recommended medications.
    There is an output sample:
    Output:
    Based on your symptoms, it sounds like you may have acute pancreatitis. To confirm this, we will need to run a series of medical tests. We will start with a blood test and a complete blood count (CBC), as well as a radiographic imaging procedure to determine the extent of the pancreatitis. We may also need to provide intravenous fluid replacement and perform kidney function tests and glucose level measurements. Additionally, a urinalysis will be necessary to check for any kidney damage.

"""
\end{lstlisting}

\end{appendix}
\end{document}